\documentclass[11pt,a4paper]{article}
\usepackage[hyperref]{naaclhlt2019}
\usepackage{times}
\usepackage{latexsym}
\usepackage{graphicx}
\usepackage{amsmath}
\usepackage[utf8]{inputenc}
\usepackage[english]{babel}
 
\usepackage{url}

\aclfinalcopy 
\def\aclpaperid{4023} 
\title{Diversifying Reply Suggestions using a Matching-Conditional Variational Autoencoder}
\begin{document}


\author{
        Budhaditya Deb \\
        \\\And
        Peter Bailey \\
        \\
        Microsoft Search, Assistance and Intelligence \\
        {\tt \{budeb, pbailey, milads\}@microsoft.com} \\\And
        Milad Shokouhi \\
  } 

\date{}
\maketitle
\begin{abstract}
	We consider the problem of diversifying automated reply suggestions for a commercial instant-messaging (IM) system (Skype). Our conversation model is a standard matching based information retrieval architecture, which consists of two parallel encoders to project messages and replies into a common feature representation. During inference, we select replies from a fixed response set using nearest neighbors in the feature space. To diversify responses, we formulate the model as a generative latent variable model with Conditional Variational Auto-Encoder (M-CVAE). We propose a constrained-sampling approach to make the variational inference in M-CVAE efficient for our production system. In offline experiments, M-CVAE consistently increased diversity by $\sim30-40\%$ without significant impact on relevance. This translated to a 5\% gain in click-rate in our online production system.
\end{abstract}

\section{Introduction}
	Automated reply suggestions or smart-replies (SR) are increasingly becoming common in many popular applications such as Gmail \citeyearpar{Kannan2016}, Skype \citeyearpar{SkypeSmartReply}, Outlook \citeyearpar{OWASmartReply}, LinkedIn \citeyearpar{LinkedInSmartReply}, and Facebook Messenger. 
	
	Given a message, the problem that SR solves is to suggest short and relevant responses that a person may select with a click to avoid any typing. For example, for a message such as \texttt{Want to meet up for lunch?} an SR system may suggest the following three responses {\{\texttt{Sure}; \texttt{No problem!};  \texttt{Ok}\}}. While these are all relevant suggestions, they are semantically equivalent.  In this paper, we consider how we can diversify the suggestions such as with {\{\texttt{Sure}; \texttt{Sorry I can't}; \texttt{What time?}\}} without losing any relevance. Our hypothesis is that encompassing greater semantic variability and intrinsic diversity will lead to higher click-rates for suggestions.

	Smart-reply has been modeled as an sequence-to-sequence (S2S) process \cite{Li2016ADO,Kannan2016,VinyalsL15} inspired by their success in machine translation. It has also been modeled as an Information Retrieval (IR) task  \cite{HendersonASSLGK17}. 
	Here, replies are selected from a fixed list of responses, using two parallel \emph{Matching} networks to encode messages and replies in a common representation. Our production system uses such a Matching architecture. 
	
	There are several practical factors in favor of the \emph{Matching-IR} approach. Production systems typically maintain a curated response-set (to have better control on the feature and to prevent inappropriate responses) due to which they rarely require a generative model. Moreover, inference is efficient in the matching architecture as vectors for the fixed response set can be pre-computed and hashed for fast lookup. Qualitatively, S2S also tends to generate generic, and sometimes incorrect responses due to label and exposure bias. Solutions for S2S during training \cite{Wiseman2016SequencetoSequenceLA} and inference \cite{Li2016ADO} have high overhead. 
	Matching architectures on the other hand, can incorporate a global normalization factor during training to mitigate this issue \cite{Sountsov2016LengthBI}. 

    \begin{figure}
    	\centering
    	\includegraphics[scale=0.45]{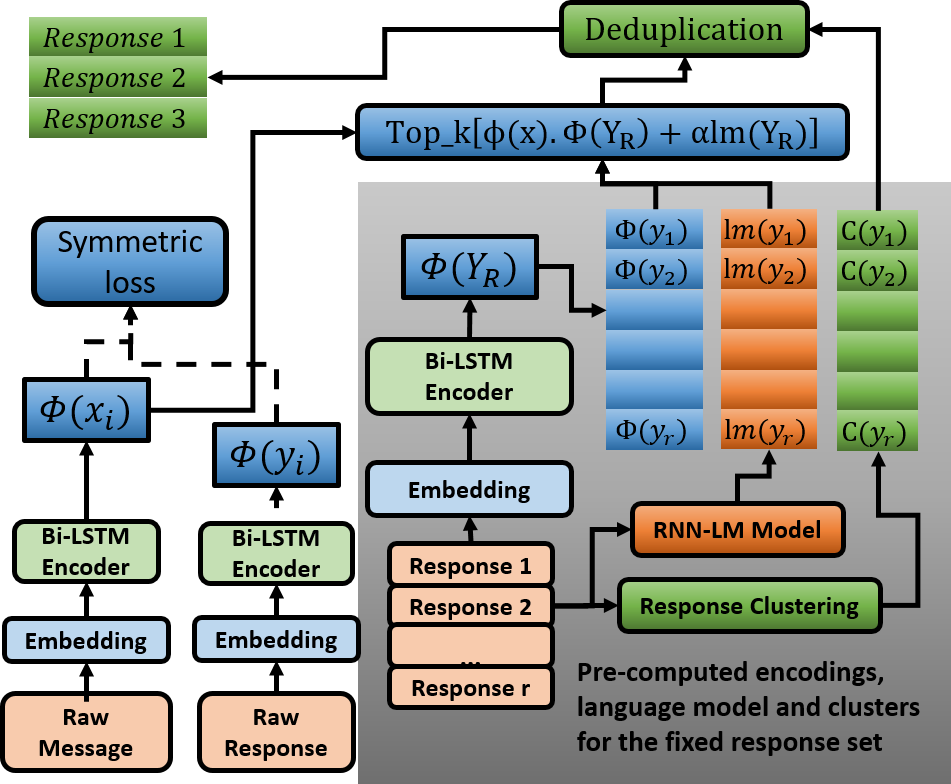}
    	\caption{Traning and inference graph for \emph{Matching}. During inference, the response side stack is pre-computed (shaded grey).}
    	\label{fig:Matching-Architecture}
    \end{figure}
	
    In practice we found that the Matching model retrieves responses which are semantically very similar in lexical content and underlying intent as shown in (Table \ref{tab:CompareDuplicates}). This behavior is not surprising and even expected since we optimize the model as a point estimation on golden message-reply (m-r) pairs. In fact, it illustrates the effectiveness of encoding similar intents in the common feature space. While this leads to individual responses being highly relevant, the model needs to diversify the responses to improve the overall relevance of the set by covering a wider variety of intents. We hypothesize that diversity would improve the click rates in our production system. This is the main focus of this paper. We provide two baselines approaches using lexical clustering and maximal marginal relevance (MMR) for diversification in the Matching model.
    

	Since we typically do not have multiple responses in one-on-one conversational data (and thus cannot train for multiple-intents), we consider a generative Latent Variable Model (LVM) to learn the hidden intents from individual m-r pairs. Our key hypothesis is that intents can be encoded through a latent variable, which can be then be utilized to generate diverse responses.

	To this end, we propose the Matching-CVAE (M-CVAE) architecture, which introduces a generative LVM on the Matching-IR model using the neural variational autoencoder (VAE) framework \cite{Kingma2013AutoEncodingVB}. M-CVAE is trained to generate the vector representation of the response conditioned on the input message and a stochastic latent variable. During inference we sample responses for a message and use voting to rank candidates. To reduce latency, we propose a constrained sampling strategy for M-CVAE which makes variational inference feasible for production systems.
	We show that the Matching architecture maintains the relevance advantages and inference-efficiency required for a production system while CVAE allows diversification of responses. 
	
	We first describe our current production model and diversification approaches. Next, we present our key contribution: Matching-CVAE. Finally we report on our results from offline and online experiments, including production system performance. 
	
  
    \begin{table}[t]
    	\centering{}
    	\includegraphics[scale=0.80]{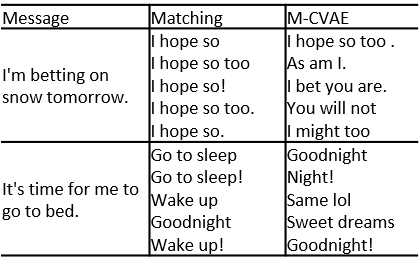}
    	\caption{The top responses (w/o de-duplication) for Matching and M-CVAE.  
    	\label{tab:CompareDuplicates}}
    \end{table}
\section{Matching Model}
	Our training data consists of message reply (m-r) pairs $[x_{i},y_{i}]$ from one-on-one IM conversations\footnote{Multi-user conversations were difficult to align reliably, given highly restricted access to preserve our users privacy.}. 
	A parallel stack of embedding and bi-directional LSTM layers encodes the raw text of m-r by concatenating the last hidden state of the backward and forward recurrences as $\Phi_{X}(x_{i})$ and $\Phi_{Y}(y_{i})$ (Figure \ref{fig:Matching-Architecture}). The encodings are trained to map to a common feature representation using the  \emph{symmetric-loss}: a probabilistic measure of the similarity as a normalized dot product $\Theta_{x_{i}y_{i}}=\Phi_X(x_{i})\cdot\Phi_Y(y_{i})$ in equation \ref{eq:symmetric_loss}. We maximize the $-\ln p(\Theta)$ during  training. 
	
	Note the denominator in the \emph{symmetric-loss} is different from a softmax (where the marginalization is usually over the $y$ terms) to approximate $p(y_{i}|x_{i})$. Instead, it the sums over each message w.r.t. all responses and vice-versa. This normalization (analogous to a Jaccard index) in both directions enforces stronger constraints for a dialog pair\footnote{\citet{Li2016ADO} made a similar argument with Mutual Information penalty during inference.}. Thus, it is more appropriate for a conversational model where the goal is \emph{conversation compatibility} rather than \textit{content similarity}. Symmetric loss improved the relevance in our model. We omit the results here, to focus on diversity.

	{\small
		\begin{gather}
			p(\Theta_{x_i y_i})=\frac{
				e^{\Theta_{x_i y_i}}}
				{{\textstyle {\sum_{y_{j}}{ e^{\Theta_{x_{i}y_{j}}}}+\sum_{x_{j}}{e^{\Theta_{x_{j}y_{i}}}}-e^{\Theta_{x_{i}y_{i}}}}}}\label{eq:symmetric_loss}\\
			S_k(x)=\textup{softmax}_k[top_k[\Phi_X(x)\cdot\Phi_Y(Y_R)+\alpha lm(Y_R)]]\label{eq:matching_score}
		\end{gather}
	}
	During inference, we pre-compute the response vectors $\Phi_{Y}(Y_{R})$ for a fixed response set $Y_{R}$. We encode an input $x$ as $\Phi_X(x)$, and find the $K$ nearest responses $Y_{R_k}$, using a score composed of the dot product of $\Phi_{X}(x)$ and $\Phi_{Y}(Y_R)$ and a language-model penalty $lm(Y_R)$\footnote{We train an LSTM language model on the training data.} in equation \ref{eq:matching_score}. The $lm(Y_R)$ is intended to suppress very specific responses similar to \cite{HendersonASSLGK17}. The $\alpha$ parameter is tuned separately on an evaluation set. We de-duplicate the $Y_{R_k}$ candidates and select top three as suggested replies. The training and inference graph is shown in Figure \ref{fig:Matching-Architecture}.
	
    \begin{figure*}[t]
    	\centering
    	\includegraphics[scale=0.40]{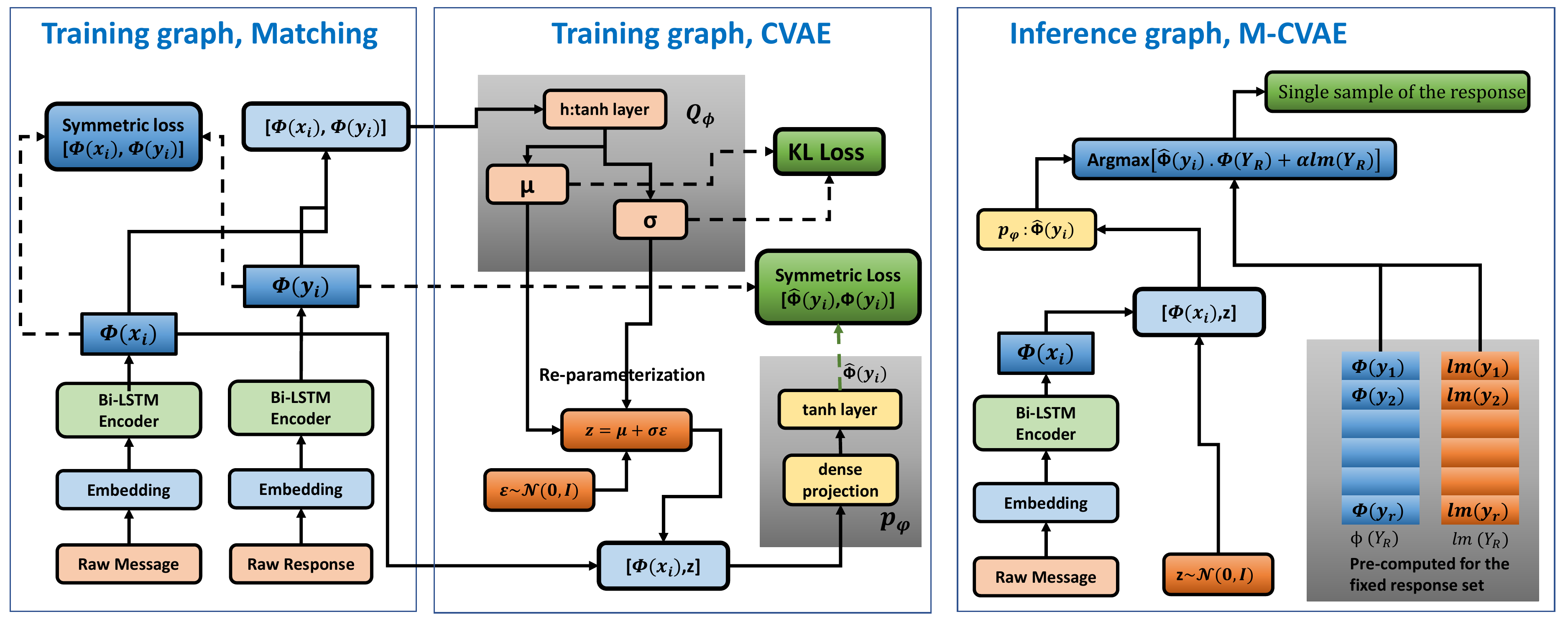}
    	\caption{Matching (left) and CVAE (center) training models. Dotted arrows show the inputs to the loss functions. Right side shows the M-CVAE inference network where the shaded region shows the pre-computed values of the fixed response set..}
    	\label{fig:Matching-CVAE-Architecture}
    \end{figure*}
    \subsection{Response Diversification}
    	The matching model by itself, retrieves very similar responses as shown in Table \ref{tab:CompareDuplicates}. Clearly, the responses need to be de-duplicated to improve the quality of suggestions. We present two baseline approaches to increase diversity.
    
        \textbf{Lexical Clustering (LC)}: 
    	Table \ref{tab:CompareDuplicates}, motivates the use of simple lexical rules for de-duplication. We cluster responses which only differ in punctuations \texttt{(Thanks!, Thanks.)}, contractions \texttt{(cannot:can't, okay:ok)}, synonyms \texttt{(yeah, yes, ya)} etc. We further refine the clusters by joining responses with one-word edit distance between them \texttt{(Thank you so much. Thank you very much)} except for negations. During inference, we de-duplicate candidates belonging to the same clusters. 
      
    
        \textbf{Maximal Marginal Relevance (MMR)}:
    	As a way to increase the diversity in IR, \cite{MMR} introduced the MMR criterion to penalize the query-document similarity with inter-document similarity to rank candidates using \textit{marginal} relevance. 
    	
    	In the context of the SR, we apply the MMR principle as follows. First, we select the $K$ candidates, (with scores $S_k(x)$ and response vectors $\Phi_Y(Y_{R_k}$)) using equation \ref{eq:matching_score}. Next, we compute the the novelty $N_k$ (or marginal relevance) of the $k^{th}$ response with respect to the other $K-1$ candidates using equation~\ref{eq:novelty}.
    	Finally, we re-rank the candidates using the MMR score computed from equation \ref{eq:mmr_score}. Our MMR implementation is an approximation of the original (which is iterative). Nevertheless, it allows the ranking in one single forward pass and thus is very efficient in terms of latency.
    
    	{\small
    			\begin{gather}
    				N_k=\frac{1}{K-1} \sum_{j\neq k}^K CosSim(\Phi_Y(Y_{R_k}),\Phi_Y(Y_{R_j})) \label{eq:novelty}\\
    				MMR_k(x) = \beta S_k(x) - (1-\beta)N_k
    				\label{eq:mmr_score}
    			\end{gather}
    	}
    	Table \ref{tab:MMR_IM} shows that LC and MMR are quite effective at reducing duplicates. We have also explored other clustering approaches using embeddings from unsupervised models, but they were not as effective as LC or MMR. 
\section{Matching-CVAE (M-CVAE)}
    Neither MMR nor LC solves the core issue with diversification i.e., learning to suggest diverse responses from individual m-r pairs. Privacy restrictions prevent any access to the underlying training data for explicit annotation and modeling for intents.  Instead, we model the hidden intents in individual m-r pairs using a latent variable model (LVM) in M-CVAE.
    
    In M-CVAE we generate a response vector conditioned on the message vector and a stochastic latent vector. The generated response vector is then used to select the corresponding raw response text. 

	M-CVAE relies on two hypotheses. First, the encoded vectors are accurate distributional indexes for raw text. Second, the latent variable encodes intents (i.e. a manifold assumption that similar intents have the same latent structure). Thus, samples from different latent vectors can be used to generate and select diverse responses within the Matching-IR framework.

	We start with a base Matching model which encodes an m-r pair as $\Phi_X(x_{i})$ and $\Phi_Y(y_{i})$. We assume a stochastic vector $z$ which encodes a latent intent, such that $\Phi_Y(y_i)$ is generated conditioned on $\Phi_X(x_i)$ and $z$. The purpose of learning the LVM is to maximize the probability of observations $\Phi_X,\Phi_Y$ by marginalizing over $z$. This is typically infeasible in a high dimensional space. 
	
	Instead, the variational framework seeks to learn a posterior $Q_{\phi}(z|\Phi_{X},\Phi_{Y})$ and a generating $p_{\varphi}\left(\Phi_{Y}|\Phi_{X},z\right)$ function to directly approximate the marginals.
	In the neural variational framework ~\cite{Kingma2013AutoEncodingVB} and the conditional variant CVAE \cite{Sohn_CVAE}, the functionals $Q_{\phi}$ and $p_{\varphi}$ are approximated using non-linear neural layers\footnote{Also referred as inference/recognition and reconstruction networks, they appear like an auto-encoder network.}, and trained using Stochastic Gradient Variational Bayes (SGVB). 

	We use two feed forward layers for $Q_{\phi}$ and $p_{\varphi}$ as shown in equations \ref{eq:CVAE_Encoder} and  \ref{eq:CVAE_Decoder}. Here, $\overleftrightarrow{}$ denotes the concatenation of two vectors. To sample from $Q_{\phi}$, we use the re-parameterization trick of Kingma~\shortcite{Kingma2013AutoEncodingVB}. First, we encode the input vectors interpreted as mean and variance $\left[\mu,\sigma^{2}\right]$. Next, we transform to a multivariate Gaussian form by sampling $\varepsilon\sim\mathcal{N}(0,I)$, and apply the linear transformation in equation \ref{eq:Reparameterization}. We reconstruct the response vector as $\hat{\Phi}_Y$ with $p_{\varphi}$ (equation \ref{eq:CVAE_Decoder}). Figure \ref{fig:Matching-CVAE-Architecture} shows the complete M-CVAE architecture.
	
	The network is trained with the evidence lower bound objective (ELBO) by conditioning the standard VAE loss with response $\Phi_Y$ in equation \ref{eq:ELBO}. The first term can be computed in closed form as it is the KL Divergence between two Normal distributions. The second term denotes the reconstruction loss for the response vector. We compute the reconstruction error using the symmetric loss, $p(\hat{\Phi}_Y(y_{i}),\Phi_{Y}(y_{i}))$ from equation \ref{eq:symmetric_loss} in the training minibatch. As is standard in SGVB, we use only one sample per item during training.
	
	{\small
		\begin{gather}
			\begin{array}{c}
				h=\tanh\left(w_{\mu_{1}}^{\phi}\cdot\overleftrightarrow{\Phi_{X}\Phi_{Y}}+b_{\mu_{1}}^{\phi}\right)\\
				\mu=w_{\mu_{2}}^{\phi}\cdot h+b_{\mu_{2}}^{\phi}\\
				\sigma=\exp\left(\left(w_{\sigma{}_{2}}^{\phi}\cdot h+b_{\sigma_{2}}^{\phi}\right)/2\right)\\ 
			\end{array}	\label{eq:CVAE_Encoder}\\
			z\sim Q_{\phi}=\mu+\sigma\cdot\varepsilon,\; where \; \varepsilon\sim\mathcal{N}(0,I)\label{eq:Reparameterization}\\
			\hat{\Phi}_Y:p_{\varphi}=w_{2}^{\varphi}\cdot\tanh\left(w_{1}^{\varphi}\cdot\overleftrightarrow{z\Phi_{X}}+b_{1}^{\varphi}\right)+b_{2}^{\varphi}\label{eq:CVAE_Decoder}\\
			ELBO=-D_{KL}\left[Q_{\phi}\left(z|\Phi_{X},\Phi_{Y}\right)\Vert p\left(z|\Phi_{X}, \Phi_{X}\right)\right]\nonumber\\
			+E\left[\ln p_{\varphi}(\Phi_{Y}|z,\Phi_{X})\right]\label{eq:ELBO}\\
			Predict:Argmax_{Y_{R}}[\hat{\Phi}_Y(y_i)\cdot\Phi_Y(Y_{R})+lm(Y_{R})]\label{eq:MCVAE_Inference}
		\end{gather}
	}
	\subsection{Inference in CVAE}
		During inference, We pre-compute the response vectors $\Phi_Y(Y_{R})$ and $lm(Y_R)$ scores as before. However, instead of matching the message vector with the response vectors, we find the nearest-neighbors of the \textit{generated} response vector, $\hat{\Phi}_Y$ with $\Phi_Y(Y_R)$ . Next, we use a sampling and voting strategy to rank the response candidates. 
		
		\textbf{Sampling Responses}: To generate $\hat{\Phi}_Y$, we first sample $z\sim\mathcal{N}(0,I)$, concatenate with $\Phi(x)$ and generate $\hat{\Phi}(y)$ with the decoder $p_{\varphi}$ from equation \ref{eq:CVAE_Decoder}. The sampling process is shown Figure \ref{fig:Matching-CVAE-Architecture} (right).
		
		\textbf{Voting Responses}: The predicted response sample for a given input and a $z$ sample is given by equation \ref{eq:MCVAE_Inference}. In each sample, a candidate response (argmax) gets the winning vote. We generate a large number of such samples and use the total votes accumulated by responses as a proxy to estimate the likelihood $p(y|x)$. Finally, we use the voting-score to rank the candidates in M-CVAE. 

    \subsection{Constrained sampling in CVAE} 
		To deploy M-CVAE in production we needed to solve two issues. First, generating a large number of samples significantly increased the latency compared to Matching. Reducing the number of samples leads to higher variance where M-CVAE can sometimes select diverse but irrelevant responses (compared to Matching which selects relevant but duplicate responses). We propose a \emph{constrained sampling} strategy which solves both these problems by allowing better trade off between diversity and relevance at a reduced cost. 

        We note that the latency bottleneck is essentially in the large dot product with pre-computed response vectors (our response set size is \textasciitilde{}30k) in equation \ref{eq:MCVAE_Inference}. Here, the number of matrix multiplications for $N$ samples is $600*30000*N$ (with encoding dimension size of 600). However, during the sampling process, only a few relevant candidates actually get a vote. Thus, we can reduce this cost by pre-selecting top $K$ candidates using the Matching score (eq. \ref{eq:matching_score}) and then pruning the response vector to the selected $K$ candidates. This \textit{constrains} the dot-product in each sampling step to only $K$ vectors, and reduces the number of matrix multiplications for $N$ samples to $600*K*N$, where $K\ll30000$.
        
        By pruning the response set, we are able to fit all the sampling vectors within a single matrix, and apply the entire sampling and voting step as matrix operations in one forward pass through the network. This leads to an extremely efficient graph and allows us to deploy the model in production.

	    \textbf{Sampling with MMR}: As seen in Table \ref{tab:CompareDuplicates}, the candidates selected using Matching score can have very low diversity to begin with and can reduce the effectiveness of M-CVAE. To diversify the initial candidates, we can use our MMR ranking approach as follows. We first select top $2K$ responses using Matching and compute the MMR scores from equation \ref{eq:mmr_score}. Next, we use the MMR scores to select the top $K$ diverse responses for use in constrained sampling in M-CVAE. 

		All the inference components (Matching, MMR, and constrained sampling), when applied together requires just one forward pass through the network. Thus, we can not only trade-off diversity and relevance, but also control the latency at the same time. Constrained sampling was critical for deploying to production systems.
		
        
\section{Experiments and Results}
    Our current production model in Skype is a parallel Matching stack (Figure \ref{fig:Matching-Architecture}) with embedding size of $320$ and 2 Bi-LSTM layers with hidden size of $300$ for both messages and replies. The token vocabulary is \textasciitilde{}100k (tokens with a minimum frequency of 50 in the training set), and the response set size is \textasciitilde{}30k. It selects top $15$ candidates and de-duplicates using lexical clustering to suggest three responses. The entire system is implemented on the Cognitive Toolkit \cite{CNTK} which provides efficient training and run-time libraries, particularly suited to RNN based architectures.	
	
	We analyze the M-CVAE model in comparison to this production model \footnote{Since the production model has gone through numerous parameter tuning and flights, we assume that to be a strong baseline to compare with.}. The production model is also used as the control for online A/B testing, so it is natural to use the same model for offline analysis.  To train the M-CVAE, we use the base Matching model, freeze its parameters, and then train the CVAE layers on top. We apply a dropout rate of 0.2 after the initial embedding layer (for both Matching and M-CVAE) and use the Adadelta learner for training. We use the loss on a held out validation set for model selection. 
 
    \textbf{Training data}: We sample \textasciitilde{100} million pairs of m-r pairs from one-on-one IM conversations. We filter out multi-user and multi-turn conversations since they were difficult to align reliably. We set aside 10\% of the data to compute validation losses for model selection. The data is completely eyes-off i.e., neither the training nor the validation set is accessible for eyes-on analysis. 

	\textbf{Response set}: To generate the response set, we filter replies from the m-r pairs with spam, offensive, and English vocabulary filters and clean them of personally identifiable information. Next, we select top 100k responses based on frequency and then top 30k based on lm-scores. We pre-compute the lm-scores, lexical-clusters and encodings for the response set and embed them inside the inference graphs as shown in Figure \ref{fig:Matching-Architecture} and \ref{fig:Matching-CVAE-Architecture}. 

    \textbf{Evaluation metrics and set}: The model predicts three responses per message for which we compute two metrics: \emph{Defects} (a response is deemed incorrect) and \emph{Duplicates} (at least 2 out of 3 responses are semantically similar). We use crowd sourced human judgments with at least 5 judges per sample. Judges are asked to provide a binary \textit{Yes/No} answer on defects and duplicates. Judge consensus (inter annotator agreement) of 4 and above is considered for metrics, with 3 deemed as \textit{no-consensus} (around 5\%). Since training/validation sets are not accessible for analysis, we created an \textit{evaluation} set of 2000 messages using crowd sourcing for reporting our metrics.
    
	
    \begin{table}[t]
    	\centering{}
    	\includegraphics[scale=0.85]{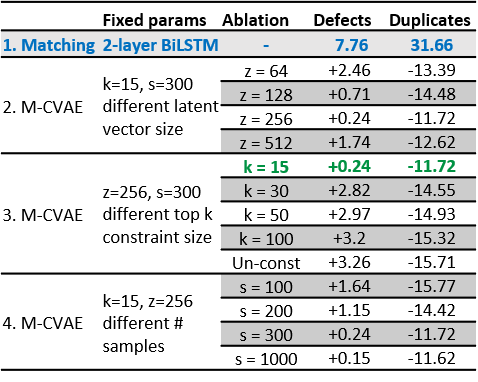}
    	\caption{Relative change in \textit{Defect} and \textit{Duplicate} metrics for different hyper-parameters of M-CVAE w.r.t. to the baseline Matching (row 1). In all cases, M-CVAE significantly reduces duplicates with minor increase in defects w.r.t. to the baseline model. Row highlighted in green is the configuration chosen for online A/B test.
    	\label{tab:M_CVAE_Ablation}}
    \end{table}
    \begin{table}[t]
    	\centering{}
    	\includegraphics[scale=0.85]{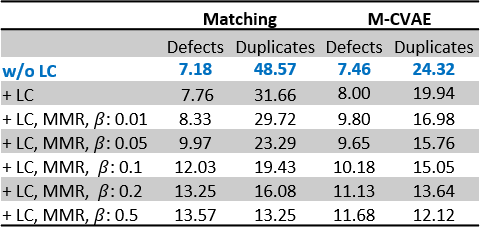}
    	\caption{Contribution of baseline diversification techniques, LC and MMR on duplicate reduction. 
    	\label{tab:MMR_IM}}
    \end{table}
    \begin{table}[t]
    	\centering{}
    	\includegraphics[scale=0.85]{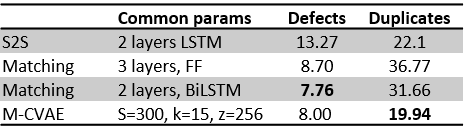}
    	\caption{Summary metrics comparing architectures. 
    	\label{tab:BaseMatchingS2S}}
    \end{table}
    
    \begin{table}[t]
    	\centering{}
    	\includegraphics[scale=0.85]{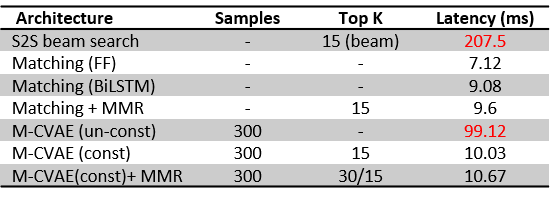}
    	\caption{Latency (in ms) in production servers\label{tab:Latency}.}
    \end{table}
    \begin{table}[t]
    	\centering{}
    	\includegraphics[scale=0.85]{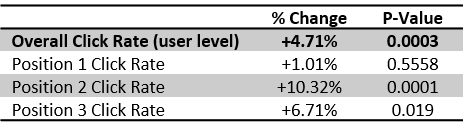}
    	\caption{Click rates for the M-CVAE flighted model. The control is the Matching model in production.\label{tab:M_CVAE_Flight_Results}}
    \end{table}
    \textbf{M-CVAE parameters}: We consider three parameters for ablation studies in M-CVAE: size of latent vector $z$, number of samples $s$ and the response pruning size $k$ for constrained sampling. The results are shown in Table \ref{tab:M_CVAE_Ablation}. The M-CVAE numbers (row 2 onwards) are relative to the base Matching model in row 1. First, row 2 shows that latent vector size of 256 provides a suitable balance between defects and duplicates, but in general, the size of latent variable is not a significant factor in performance. Next, in row 3, we see that the response-pruning size $k$, is an effective control to trade-off defects and duplicates. Thus constrained sampling not only reduces the latency but also provides quality control required in a production system. In row 4, we see that more samples lead to better metrics but the improvements are marginal beyond 300 samples. In all cases, M-CVAE significantly reduces duplicates (by as much as 40\%) without any major increase in defects. We select the model with hyper-parameters $[k=15, z=256, s=300]$ for further analysis.
    
	\textbf{Diversification with LC}:  The first two rows of Table \ref{tab:MMR_IM} analyzes the impact of LC based de-duplication. LC can significantly reduce the duplicates in the base matching model. However, M-CVAE (even without LC) reduces the rates by almost 50\% as shown in column 4 in row 1. Using LC as a post processing step after M-CVAE, can give further boosts in diversity (row 2).

	\textbf{Diversification with MMR}: Table \ref{tab:MMR_IM} also reports the impact of MMR re-ranking. For Matching+MMR, duplicates can reduce significantly as we increase the $\beta$ parameter, but at the cost of increased defects. With MMR+M-CVAE, further diversification can be achieved, and typically at a lower defect rate. This shows the advantage of using M-CVAE which conditions the responses on the message and hence has stronger controls on the relevance than MMR.
	
	\textbf{Comparison with other architectures}: We have considered two other architectures for our SR system. First is a standard S2S with attention \cite{DBLP:BahdanauCB14} with equivalent parameters for embedding and LSTMs as our base model, and inference using beam search decoding with width 15. Second, is a feed-forward (instead of an LSTM) based Matching encoder architecture which is equivalent to the one in \cite{HendersonASSLGK17}. All models use LC for de-duplication after 15 candidate responses are selected. Table \ref{tab:BaseMatchingS2S} validates our architectural preference towards Matching/Bi-LSTM which has a superior performance in terms of defects.
	
	
	\textbf{Inference latency}: Architecture choices were also driven by latency requirements in our production system. The results are summarized in Table \ref{tab:Latency} for different architectures. S2S and unconstrained sampling in M-CVAE were unsuitable for production due to their high latencies. With constrained sampling (including MMR), the latency increases marginally compared to the base model, and allows us to put the model in production.
	
	\textbf{Online experiments}: Offline metrics were used principally for selecting the best candidate models for online A/B experiments. We selected M-CVAE model with parameters [z=256, k=15, s=300] from Table \ref{tab:M_CVAE_Ablation}. Using our existing production model as the control, and a treatment group consisting of 10\% of our IM client users (with the same population properties as the control), we conducted an online A/B test for two weeks. Table \ref{tab:M_CVAE_Flight_Results} shows that the click-rate for M-CVAE compared to the Matching model increased by \textasciitilde{}5\% overall. 
	
	Gains were driven by the increase in the 2nd (10.3\%) and 3rd (6.7\%) suggested reply positions with virtually no impact in the 1st position. This correlates with our offline analysis since M-CVAE typically differs from the base model at these two positions. Intuitively, the three positions point to the \textit{head, torso and tail} intents of responses \footnote{which was validated by the absolute click rates for each of these positions but not shown in the table}. Gains at these positions show that M-CVAE extracts diverse responses without sacrificing the relevance of these tail intents. 
	
	Driven by these gains, we have switched our production system in Skype to use M-CVAE for 100\% of users.

\section{Related work}
    
	Several researchers have used CVAEs \cite{Sohn_CVAE} for generating text \cite{pmlr-v48-miao16, Guu2017GeneratingSB, SamBow16}, modeling conversations \cite{VHCR}, diversifying responses in dialogues \cite{Zhao2017LearningDD,Shen2017ACV} and improving translations \cite{StochDecNMT}. 
	These papers use S2S architectures which we found impractical for production. We demonstrate similar objectives without having to rely on any sequential generative process, in an IR setting.
    
    VAE has been also used in IR \cite{ChaidaroonSIGIR17} to generate hash maps for semantically similar documents and top-n recommendation systems \cite{VAETopN}. In contrast, we demonstrate semantic-diversity in intents in a conversational IR model with M-CVAE.

	Novelty and diversity are well-studied problems in IR \cite{SVMDiversity, NovDivClarke2008} where it is assumed that document topics are available (and not latent) during training. Diversification effect as shown in \cite{Chen_Less_is_More} relies on relevance (click) data, and thus is not directly applicable in our system.  MMR ~\cite{MMR} is a relevant prior work which we use as a baseline. 

\section{Conclusions}
    We formulate the IR-based conversational model as a generative LVM, optimized with the CVAE framework. M-CVAE learns to diversify responses from single m-r pairs without any supervision. Online results show that diversity increases the click rates in our system. Using efficient constrained sampling approach, we have successfully shipped the M-CVAE to production. 
    
    Increase in click rates over millions of users is incredibly hard. We have also experimented with the M-CVAE model trained for suggesting replies to emails in Outlook Web App (significantly different characteristics than IM) and seen similar gains. The results across domains suggests strong generalization properties of the M-CVAE model and validates our hypothesis that increased diversity leads to higher click-rates by encompassing greater semantic variability of intents.
    
   
    Perhaps the most important quality of the M-CVAE is that response vector can be flexibly conditioned on the input and thus a transduction process. In contrast, in the Matching IR model, response vectors are pre-computed and independent of the input. M-CVAE thus opens up new avenues to improve the quality of responses further through personalization and stylization. This is the subject of future work.
    
\section*{Acknowledgments}
We gratefully acknowledge the contributions of Lei Cui, Shashank Jain, Pankaj Gulhane and Naman Mody in different parts of our production system on which this work builds upon. We also thank Chris Quirk and Kieran McDonald for their insightful feedback during the initial development of this work. Finally, we thank our partner teams (Skype, infrastructure, and online experimentation) for their support. 
\bibliography{naacl2019}
\bibliographystyle{acl_natbib}
\end{document}